\documentclass[10pt,twocolumn,letterpaper]{article}

\usepackage{cvpr}      

\definecolor{cvprblue}{rgb}{0.21,0.49,0.74}
\usepackage[pagebackref,breaklinks,colorlinks,allcolors=cvprblue]{hyperref}

\def\paperID{*****} 
\def\confName{CVPR}
\def\confYear{2025}

\title{Dynamic LSTM-based Memory Encoder For Long-term LLM Interactions}

\author{Evan Lou\\
University of Washington\\
{\tt\small evanlou@cs.washington.edu}
\and
Charles Li\\
University of Washington\\
{\tt\small czli2106@cs.washington.edu}
}

\begin{document}
\maketitle
\begin{abstract}

Memory storage for Large Language models (LLMs) is becoming an increasingly active area of research, particularly for enabling personalization across long conversations. We propose Pref-LSTM, a dynamic and lightweight framework that combines a BERT-based classifier with a LSTM memory module that generates memory embedding which then is soft-prompt injected into a frozen LLM. We synthetically curate a dataset of preference and non-preference conversation turns to train our BERT-based classifier. Although our LSTM-based memory encoder did not yield strong results, we find that the BERT-based classifier performs reliably in identifying explicit and implicit user preferences. Our research demonstrates the viability of using preference filtering with LSTM gating principals as an efficient path towards scalable user preference modeling, without extensive overhead and fine-tuning.
\end{abstract}

\section{Introduction}
\label{sec:intro}
\raggedbottom

\paragraph{}
Large Language Models (LLMs) have demonstrated remarkable capabilities across a wide range of natural language processing tasks. This includes excelling at open-domain dialogue, question answering, writing, and task assistance. Through exhaustive and extensive training, these models have learned how to generalize across a myriad of domains and perform at an extremely high level. However, despite these successes, LLMs are limited in their ability to persistently remember and incorporate user-specific preferences in a long-term context. 

As LLM use expands to personal assistants, educational instructors, and domain-specific agents, the topic of LLM memory becomes more and more imminent. In order to provide coherent and personalized interactions with users, it is essential that LLMs can maintain a running memory of user preferences that dynamically update. With dynamic memory state, models can learn to optimize their responses tailored towards any user they are “conversing” with, enabling models to scale personalization effectively. For example, if a user mentions in a prompt \textit{``I’m lactose intolerant"}, and then in a later prompt asks for \textit{``dessert recommendations,"} the model should learn to avoid suggesting foods that contain high amounts of dairy.

\begin{figure}[hb]
  \centering
  \includegraphics[width=0.8\linewidth]{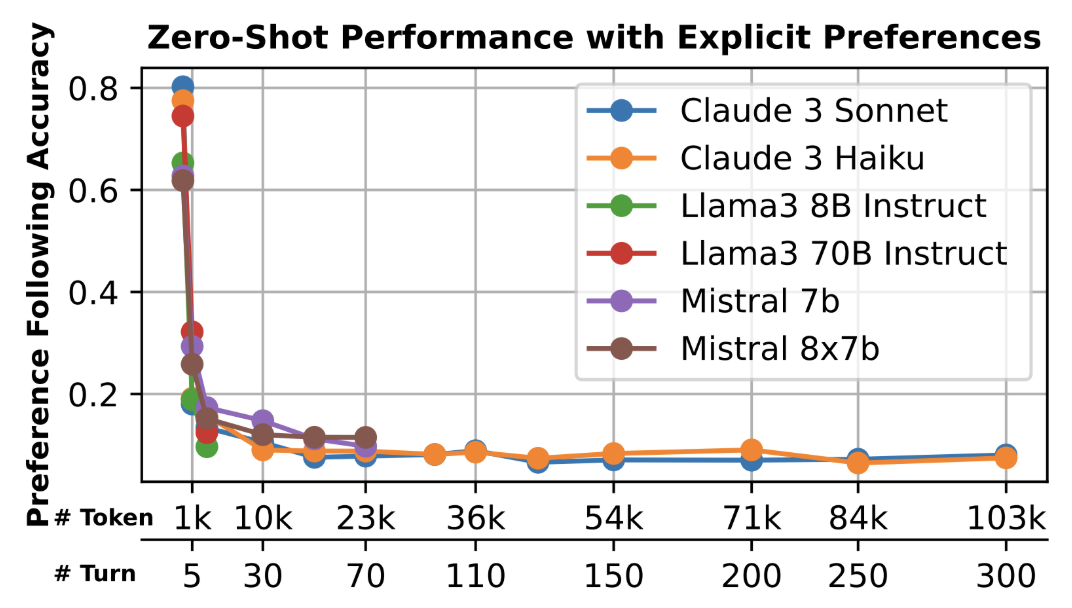}
  \caption{Pref-Eval Benchmark on state of the art LLMs}
  \label{fig:example}
\end{figure}

Effective personalization requires robust LLM understanding of implicit and explicit user behavior signals. Although users may explicitly state their likes and dislikes by mentioning phrases like \textit{``I don’t like cats,"} these preferences can also be learned implicitly through indirect cues such as positive/negative responses to LLM answers and subtle stylistic changes in user behavior. The challenge not only lies in discerning preferences, but also in deciding which preferences to keep, and which ones to discard in the case of conflicting preferences. Retaining too much information can lead to noisy memory representations, inefficient memory retrieval, and LLM hallucinations, while too little information storage can lead to compromised performance. Furthermore, as context windows become larger and larger, the issue of forgetting information is extenuated.

\subsection{Related Works}
\paragraph{LLM Personalization and Prompt Embedding}
Early LLM personalization studied the model's ability to follow user personality based on fine-tuned examples~\cite{zhang2018neural}. These methods inherently teach models to follow user personality through a multi-interaction conversation within the same token window. Attempts at prompt injection~\cite{mazare2018training} using word tokens that represent user profiles reflect the need to remind LLMs of previous context in future prompts. RevGAN~\cite{li2019revgan} proposes injecting vector embeddings of items and users into LLMs, benefiting from more efficient representations of preferences in latent space. UserLLM~\cite{lin2024userllm} extends this by integrating preference embeddings into LLMs by fusing cross-attention and latent feature representation. In addition to employing cross-attention, techniques such as soft-prompting have evolved as a way to inject preference embeddings~\cite{doddapaneni2024softprompt, liu2023llava}, but these techniques remain limited in adaptiveness across time and lack temporal understanding.
\paragraph{External Memory-Storage and Extended Context Windows}
Methods like LaMP and MemoRAG~\cite{salemi2023lamp, wang2024rolellm, longcontext2024, wei2023flan} focus on explicit data storage and retrieval methods of preference data stored as memory. These techniques do dynamically update memory storage but are computationally intensive and could benefit from more efficient latent memory representations. Other strategies include expanding context window length~\cite{longcontext2024} and fine tuning models to follow preferences better in a single context window by learning to “remember” previous context~\cite{zhao2024prefeval, wei2023flan}. However, these methods focus on improving model preference following in a single context window, whereas our method emphasizes dynamic memory storage, potentially eliminating the need for LLMs to focus on prior context.

\paragraph{Context Compression and Recurrent Transformer} Compressed Context Memory~\cite{kim2023compressed} dynamically compresses key/value pairs during online interactions, enabling continual dialogue without overwhelming memory constraints. Similarly, recurrent transformer architectures~\cite{bulatov2022recurrent} enhance long-term personalization by allowing models to pass information segments through memory tokens and associative memory networks. These approaches focus on retaining user-specific information in compact, learnable memory modules. However, context compression and recurrent memorization for LLMs is computationally expensive and designed for general history retention, which further inspired us to create a lightweight, user-preference focused memory controller for LLMs.

\subsection{Proposed Solution}
\paragraph{}

Existing approaches mainly focus on augmenting prompts either through summaries of contextual information and using predefined representation of user-preferences that are either prompt injected or retrieved using techniques like Retrieval-Augmented Generation (RAG)~\cite{lewis2020retrieval}. However, these approaches are constrained by the context size length of models and induce significant computational overhead, reducing inference efficiency.

In this work, we present a dynamic light-weight preference-aware memory system that combines efficient and intelligent preference selection with dynamic memory embeddings. Inspired by the LSTM’s gating principles~\cite{hochreiter1997long} that enables selective retention, we design a pipeline that first observes context to decide relevant parts of past user utterances that indicate preferences, and then encodes those preferences into preference embeddings that update our internal memory representation through an LSTM. Then, the internal memory representation will be projected into word-embedding space and soft-prompted into an LLM. Inspired by how humans remember information, our memorization module will decide what part of user inputs are indicative of use preferences, and then our embedding model will convert these preferences into a ``memory” state that is most efficient and easy to retrieve from.

\par
\vspace{1em}
\noindent\textbf{Our contributions are twofold:}
\vspace{1em}
\begin{itemize}
  \item We propose a dynamic, LSTM-inspired memory storage system that continuously evolves with more context.
  \vspace{1em}
  \item We evaluate our approach against baseline model performance for preference following, and compare it to other preference-following approaches.
\end{itemize}
\vspace{1em}

We hope to not only highlight the importance of storing memory, but to introduce a new way of thinking of how to store the right memory. Memory—the ability to know what to store, efficiently store, and quickly retrieval—is an essential characteristic of humans, and by giving LLMs this ability, it has the potential to significantly boost LLM performance.

\section{Methodology}
\label{sec:methodology}

In this section, we introduce the Pref-LSTM framework that contextualizes LLMs with memory embeddings. Specifically, our framework selects preference-based dialogues from user input, encodes the dialogues into preference embeddings, and updates its internal memory recurrently with new preferences detected.

\begin{figure}[hb]
  \centering
  \includegraphics[width=0.9\linewidth]{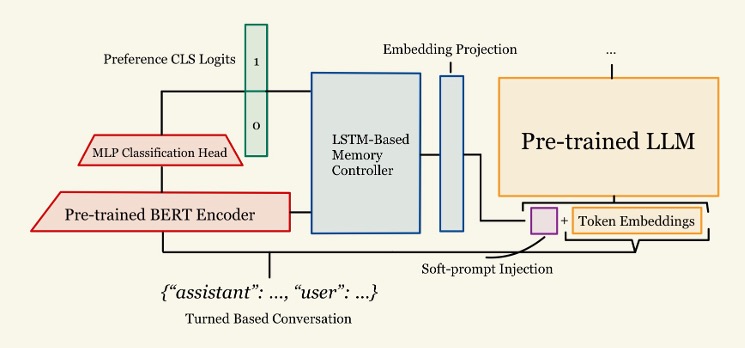}
  \caption{Pref-LSTM Two Phase Architecture}
  \label{fig:example1}
\end{figure}

\subsection{Problem Statement}
Given a user $\mathcal{U}$ and a set of its past inputs $\mathcal{V} = \{v_1, v_2, ..., v_{t - 1}, v{t}\}$, where $v_i$ is the input in timestep $i$, we can model the preference indicator of the user as a subset of its inputs $\mathcal{P} \subseteq \mathcal{V}$ such that every distinct input $p \in \mathcal{P}$ describes the user's preference on a certain subject. Consequently, every input $x \in \mathcal{V} \setminus \mathcal{S}$ does not contribute to creating user preference contexts and should be ignored by our preference memory controller. Pref-LSTM's goal is to recognize and extract the set of preference indicators $\mathcal{S}$ from the inputs $\mathcal{V}$ and contextualize the indicators to be injected into a frozen LLM to produce user-preference based responses.

\subsection{Pref-LSTM}
Pref-LSTM framework can be separated into two primary components: \textbf{memorization} and \textbf{inference}. During \textbf{memorization}, the internal memory embedding is updated based on the preference detection of the incoming user input. During \textbf{inference}, the memory embedding is fed into the downstream LLM to assist it in generating personalized responses. 

\subsubsection{Preference Detection}
Pref-LSTM explores two approaches for detecting user preferences from incoming conversation turn, which contains a LLM output and a user response.

\paragraph{BERT-Based Text Classification}
To identify whether a user utterance contains a preference, we train an MLP layer~\cite{rumelhart1986learning} with softmax on top of a frozen pretrained BERT model~\cite{devlin2019bert} as a binary classifier. Specifically, we frame preference detection as a supervised classification task: given an input sequence \( x = (x_1, x_2, \ldots, x_n) \), the model predicts a binary label \( y \in \{0, 1\} \), where \( y = 1 \) denotes a preference-related utterance and \( y = 0 \) indicates a non-preference utterance.

We use the [CLS] token representation from BERT as a summary of the input sequence. This representation is passed through the pretrained BERT, which then feeds rich features into a linear classification head followed by a softmax layer to produce probability scores over the two classes. The model is trained using standard cross-entropy loss~\cite{kullback1951on} on a dataset annotated with binary preference labels. In practice, this enables the system to filter out non-preference utterances and only update the memory module when a preference is detected.

\paragraph{Heuristic-Based Semantic Analysis}
As a lightweight alternative to model-based classification, Pref-LSTM also implements a rule-based preference detection module grounded in semantic heuristics. This method identifies preference-bearing utterances through pattern recognition, and keyword matching.

Specifically, a user utterance is classified as expressing a preference if it contains first-person subject phrases (e.g., “I like”, “I prefer”, “I usually”, “I hate”) in conjunction with opinionated adjectives or nouns (e.g., “spicy”, “quiet”, “cheap”, “movies”). To improve precision, the system leverages part-of-speech tagging and dependency parsing to verify that such phrases involve a first-person subject acting on or describing an object.

This rule-based approach does not require labeled training data and offers interpretable behavior, making it suitable for rapid prototyping and low-resource deployments. However, it may lack the flexibility and generalization capacity of learned models in handling subtle or domain-specific expressions of user preference.

\subsubsection{Memory Embedding Update}
Based on the classification result via preference detection, Pref-LSTM will filter out non-preference conversation turns. If Pref-LSTM detects a preference, it will then embed the incoming preference and combine it with the existing memory embedding.

\paragraph{User Preference Encoding} 
If the incoming conversation turn is classified as preference-indicating, the LLM-user utterance—represented as a token sequence $\{t_1, t_2, ..., t_k\}$ of length $k$—is passed through a lightweight transformer-based encoder such as MiniLM. The encoder outputs a fixed-length semantic embedding $\mathbf{e} = [e_1, e_2, ..., e_l] \in \mathbb{R}^l$ that captures the contextual meaning of the preference expressed in the input.

This embedding acts as a dense summary of the user’s stated preference, incorporating both syntactic and semantic information learned from pretraining.

The resulting preference embedding $\mathbf{e}$ is then passed into the memory controller module of Pref-LSTM, which integrates it with the current memory state.

\paragraph{Memory State Update}
At each timestep \( t \) during user-LLM interaction, the current memory state \( M_t \) in Pref-LSTM is computed as an update function over the previous memory state \( M_{t-1} \) and the preference classification result for the input utterance \( x_t \):
\[
M_t = \text{Update}(M_{t-1}, \text{pred}(x_t))
\]

Specifically, if the input is classified as \emph{non-preference-indicating}, i.e.,
\[
\text{pred}(x_t) = \texttt{non-preference},
\]
then the memory remains unchanged:
\[
M_t = M_{t-1}.
\]

On the other hand, if the utterance is classified as \emph{preference-indicating}, i.e.,
\[
\text{pred}(x_t) = \texttt{preference},
\]
the memory is updated in an LSTM-inspired manner:
\[
M_t = f_t \odot M_{t-1} + (1 - f_t) \odot \bar{E}_t,
\]
where:
- \( \bar{E}_t \in \mathbb{R}^d \) is the projected preference embedding derived from the input \( x_t \),
- \( f_t \in \mathbb{R}^d \) is a learnable, element-wise **forget gate** computed as:
\[
f_t = \sigma(W_{MM} M_{t-1} + W_{EM} \bar{E}_t + b),
\]
with \( W_{MM}, W_{EM} \in \mathbb{R}^{d \times d} \) and \( b \in \mathbb{R}^d \),
- and \( \odot \) denotes element-wise multiplication.

This update mechanism allows the memory to retain or overwrite specific dimensions of the stored user preferences depending on the gating values learned during training. The updated memory embedding is then contextualized in the downstream LLM to generate personalized response.

\subsubsection{Contextualization with Memory Embedding}
To integrate long-term user preference information into the language model, Pref-LSTM leverages the current memory embedding \( M_t \) as a \textit{soft prompt} to guide generation. Since the memory embedding lives in its own latent space, we first apply a linear transformation to project it into the same embedding space as the language model's token embeddings.

Specifically, a trainable projection matrix \( W_M \in \mathbb{R}^{d_m \times d_e} \) is used to convert the memory vector \( M_t \in \mathbb{R}^{d_m} \) into a pseudo-token embedding \( T_t \in \mathbb{R}^{d_e} \), where \( d_e \) is the dimensionality of the language model's word embeddings:
\[
T_t = M_t W_M.
\]

The resulting vector \( T_t \) is prepended to the embedded user input sequence, effectively serving as a soft prompt that injects user preference information directly into the input context of the downstream language model. Because \( T_t \) resides in the same space as token embeddings, it can be seamlessly integrated without requiring architectural modifications to the language model.

This strategy enables personalized generation conditioned on user preference memory and remains compatible with both frozen and fine-tuned language models.

\subsection{Training Framework}
To effectively extract and memorize information from user expressions, we propose a two-stage training method: preference classifier training and LSTM memory training. 

\subsubsection{Preference Classifier Training}
Due to the limited amount of reliable data for training our preference classifier, we decided to synthesize our own dataset with LLM-assisted prompt generation and human filtering. 

\paragraph{Dataset Generation} Specifically, we asked our LLM agent to generate single conversation turns between an agent and a user, following prompt engineering templates. For instance, the template below instructs the LLM agent to generate a single turn between agent and user where agent explicitly asks the user about their preferences:
\begin{verbatim}
Generate a quick preference statement.
Line 1 | Agent: asks what the user’s
favorite thing about {topic} is.

Line 2 | User: gives a short, direct
answer that reveals their preference.

Two lines only: "Agent:" then "User:"...
\end{verbatim}
Here is a template that instructs the LLM agent to generate a turn where the user implicitly reveals their preference to the agent:
\begin{verbatim}
Generate a user sharing a complaint.

Line 1 — Agent: acknowledges or
to a common frustration about {topic}.

Line 2 — User: vents about their
specific problem, revealing what they 
value or dislike.

Two lines only: \"Agent:\" then
\"User:\". User replies briefly but
reveals a clear preference.
\end{verbatim}
We also created specific templates that simulate daily conversations so that we have data that is not indicative of preference.
\begin{verbatim}
Write two lines defining {topic}.

Line1 — Agent: asks “Can you define
{topic}?”

Line2 — User: delivers a clear,
dictionary‑style definition (20‑35
words) with no personal views.

Must be two lines, "Agent:" then
"User:".
\end{verbatim}
Furthermore, to ensure that our LLM agent synthesized with as much variability as possible, we hand designed variations of the above templates and curated a list of 90 topics for the model to base their conversation around. We generated a dataset of 8452 single turn conversations, each with a binary label indicating whether it is a preference or not. There were 4915 non-preference turns and 3537 preference turns.

\paragraph{Classifier Training} Given a batch of $N$ labeled utterances $\{(x_i, y_i)\}_{i=1}^N$, where $x_i$ is the input text and $y_i \in \{0, 1\}$ is the binary preference label, we pass $x_i$ through a pretrained \texttt{prajjwal1/bert-medium} BERT encoder followed by a two-layer MLP classifier to produce logits $s_i$:
\[
s_i = \mathrm{MLP}(\mathrm{BERT}(x_i))
\]
We then compute the binary cross-entropy loss:
\[
\mathcal{L} = -\frac{1}{N} \sum_{i=1}^N \left[ y_i \log(\sigma(s_i)) + (1 - y_i) \log(1 - \sigma(s_i)) \right]
\]
where $\sigma(\cdot)$ is the sigmoid activation applied to the logit $s_i$. The model is trained to minimize $\mathcal{L}$.

\subsubsection{LSTM Memory Controller Training}

To train the LSTM-based memory controller, we freeze both the upstream BERT-based classifier and the downstream LLM. This ensures that only the LSTM parameters and the embedding projection layer are updated during training. The memory controller is designed to consume a sequence of classifier outputs and update a hidden memory state that captures user-specific preference patterns over time.

We use the OASST1 dataset, which contains multi-turn assistant-user conversations with rich preference signals. Each classifier output is first projected into a shared embedding space via a linear transformation, which is learned jointly with the LSTM parameters. The projected embeddings are then fed sequentially into the LSTM controller.

At each step, the LSTM outputs a preference representation that is decoded into one of several predefined preference categories using a softmax layer. We train the controller using a cross-entropy loss between the predicted preference class and the annotated target label:
\[
\mathcal{L}_{\text{CE}} = -\sum_{i=1}^N \log p_{\theta}(y_i \mid h_i)
\]
where \( y_i \) is the ground truth preference label at step \( i \), \( h_i \) is the LSTM's hidden state, and \( p_{\theta} \) denotes the predicted class probability from the softmax head. The objective encourages the LSTM to maintain a dynamic memory of user preference signals across dialogue turns.

\section{Experiments}

We evaluate Pref-LSTM on two main aspects: the performance of the BERT-based preference classifier, and the effectiveness of the LSTM-based memory controller in assisting user-preference-aware prompt generation for language models.

\subsection{Experimental Setup}

\paragraph{Classifier Evaluation}
To assess the BERT-based preference classifier, we curated and manually annotated a set of 80 dialogue examples covering a diverse range of conversational scenarios. These handcrafted examples were used as a testbed to evaluate the classifier's ability to detect long-term user preferences.

\paragraph{Memory Controller Evaluation}
For the LSTM-based memory controller, we bench marked its performance using the \textsc{PrefEval} dataset. This evaluation focuses on the model's ability to maintain and utilize dynamic preference representations to condition downstream LLM prompts more effectively.

\subsection{Results}

\paragraph{Classifier Performance}

\begin{table}[t]
\centering
\begin{tabular}{lccc}
\toprule
\textbf{Split} & \textbf{Train} & \textbf{Val} & \textbf{Test} \\
\midrule
Accuracy & 0.95 & 0.94 & 0.90 \\
\bottomrule
\end{tabular}
\vspace{0.5em}
\caption{Classifier accuracy on custom preference classification dataset}
\label{tab:classifier_results}
\end{table}

\begin{table}[t]
\centering
\begin{tabular}{c|cc|cc}
\toprule
\textbf{Epoch} & \textbf{Train Loss} & \textbf{Train Acc} & \textbf{Val Loss} & \textbf{Val Acc} \\
\midrule
1  & 1.2758 & 0.0258 & 1.3752 & 0.0093 \\
2  & 1.2154 & 0.0291 & 1.4032 & 0.0136 \\
3  & 1.1980 & 0.0320 & 1.3814 & 0.0106 \\
4  & 1.1829 & 0.0341 & 1.4122 & 0.0106 \\
5  & 1.1678 & 0.0356 & 1.4250 & 0.0108 \\
6  & 1.1578 & 0.0360 & 1.4223 & 0.0111 \\
7  & 1.1462 & 0.0377 & 1.4264 & 0.0109 \\
8  & 1.1390 & 0.0393 & 1.4245 & 0.0126 \\
9  & 1.1292 & 0.0401 & 1.4347 & 0.0119 \\
10 & 1.1223 & 0.0405 & 1.4698 & 0.0113 \\
\bottomrule
\end{tabular}
\vspace{0.5em}
\caption{Training and validation loss and accuracy at each epoch for LSTM-based memory controller training.}
\label{tab:lstm_training_results}
\end{table}

After training for 50 epochs, our BERT-based preference classifier achieved an accuracy of 98\% on the custom test set. This strong performance can be attributed to both the expressive feature representations provided by the pretrained BERT encoder and the simplistic nature of binary classification tasks. However, we suspect that the inherent lack of variation and potential polarization in our LLM-generated dataset may have contributed to this performance.

\begin{figure}[ht]
  \centering
  \includegraphics[width=0.9\linewidth]{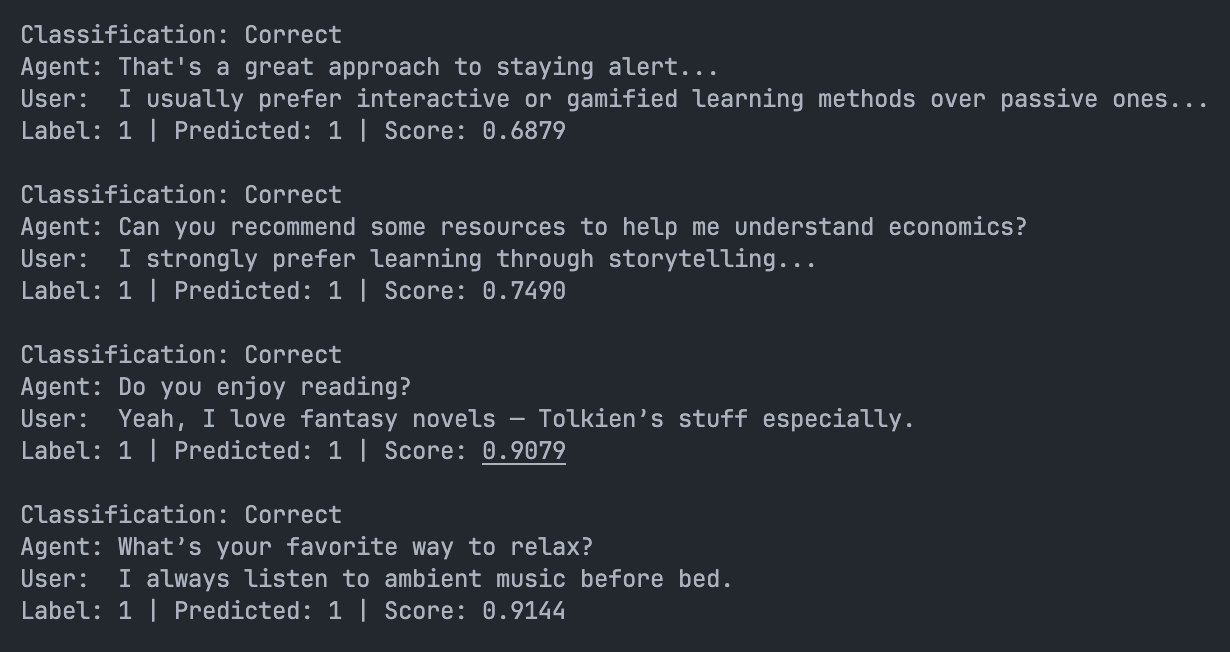}
  \caption{BERT-based Classifier Performance on Formal Language}
  \label{fig:example2}
\end{figure}

To better evaluate the models performance, we validated the classifier against a set of human handcrafted examples designed to cover a broader range of conversational styles. As illustrated in Fig.3, the classifier performs consistently on formally structured utterances that have clear grammatical structure and contain keywords such as \textit{"prefer"} or \textit{"like"}. In these cases, the classifier correctly identifies whether the user response indicates a long-term preference.

However, performance drops significantly when the conversation adopts a more casual tone as shown in Fig. 4, with less rigid grammar and informal slang. For instance, the model failed to recognize a user's strong negative sentiment toward Liverpool as a preference-indicating statement, classifying it as non-preferential. We theorize that this limitation is the result of the lack of diversity in the training data, which impacts the model’s ability to generalize to everyday conversations.
\begin{figure}[ht]
  \centering
  \includegraphics[width=0.8\linewidth]{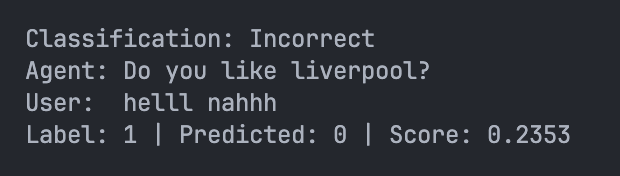}
  \caption{BERT-based Classifier Performance on Casual Language}
  \label{fig:example3}
\end{figure}

As mentioned in section 3, we also implemented a heuristic based approach to preference classification. In practice, we found this approach performed worse as it heavily influenced by word choice, and could not generalize to implicit preferences. Overall, the heuristic base approach performed worse than our trained BERT-based classifier.

\paragraph{LSTM Memory Controller Performance}
We trained our LSTM-based Memory controller using a single A100 GPU hosted by Google Colab with the downstream LLM for 10 epochs. Throughout the 10 epochs, we saw a steady decrease in training loss and an increase in validation accuracy. However, due to our resource limitations and the large size of OSAAS1, we stopped training after 10 epochs.

We tested Pref-LSTM's performance against PrefEval, the preference based LLM benchmark. Specifically, we tested the model on curated conversations where for every 3, 5, and 10 turns, a preference indicator was augmented into the dataset. In order to test the model's ability to recognize and memorize preference turns, we asked Pref-LSTM a user preference related query and then manually evaluated Pref-LSTM's response.

After extensive testing, we did not observe any improvements in preference following. Thus, we conclude that our LSTM memory module did not learn how to dynamically update it's memory embedding to assist the downstream LLM in generating preference related responses. We suppose that this is a result of many different factors: (1) The lack of compute to further our training prevented the model to learn and adapt to the downstream LLM. Most of the time, the soft word embedding that is generated by our memory controller resulted in noise, causing the LLM to produce a <end> token immediately. (2) The OSAAS1 dataset that we trained our memory controller on is not inherently a dataset that contains a lot of preference related queries. Although the OSAAS1 dataset does include some multi-turn preference conversations, a dataset that is specifically catered towards preference following would have sped up training. (3) The decision to inject user embeddings as soft-prompts may not be the best approach. While the memory embeddings themselves contain an amalgamation of previous preference, it still relies on the LLM to decide what memory embeddings to look at when generating context, which is a difficult task. (4) Currently, Pref-LSTM injects memory embeddings, regardless of user input. If memory embeddings are injected into an LLM when the generation task is not preference related, the memory embeddings could be disruptive as they would just be extra noise for the LLM's input, which could potentially interfere with LLM's performance.
\section{Conclusion and Future Work}

In this paper, we introduced Pref-LSTM, a lightweight architecture designed to aid LLM ability to follow user preferences. By filtering out context with a Bert-based text classifier and embedding user preferences into latent space via a LSTM-based Memory Controller, we a novel setup that reduces context overhead and eliminates the need for LLM fine-tuning, enabling a more scalable approach to preference following.

Our findings highlight both the promise and difficulty in building a scalable, lightweight, dynamic memory system. Although we did identify fallbacks in our synthesized dataset that led to undesirable performance in some scenarios, we demonstrate the potential and robustness of a BERT-based preference classifier and its promise in LLM memory systems. While our LSTM-based Memory controller observed no improvements in preference following, we pointed out potential future areas of improvement that could lead to superior results.

Future work includes optimizing and increasing the size of our custom preference classification dataset by including conversations from a wider variety of sentiments. Additionally, by researching new forms of communication between the LSTM memory embeddings and downstream LLM agents beyond simple soft-prompt injection has the potential to unlock the full dynamic and cost-efficiency of LSTM architectures.



\small
\clearpage
\begin{thebibliography}{18}
\providecommand{\natexlab}[1]{#1}
\providecommand{\url}[1]{\texttt{#1}}
\expandafter\ifx\csname urlstyle\endcsname\relax
  \providecommand{\doi}[1]{doi: #1}\else
  \providecommand{\doi}{doi: \begingroup \urlstyle{rm}\Url}\fi

\bibitem[Anonymous(2024)]{longcontext2024}
Anonymous.
\newblock Long context vs. rag for llms: An evaluation and revisits, 2024.
\newblock arXiv preprint arXiv:2403.00011.

\bibitem[Bulatov et~al.(2022)Bulatov, Kuratov, and Burtsev]{bulatov2022recurrent}
Aydar Bulatov, Yuri Kuratov, and Mikhail~S. Burtsev.
\newblock Recurrent memory transformer.
\newblock In \emph{Advances in Neural Information Processing Systems (NeurIPS)}, pages 11079--11091, 2022.

\bibitem[Devlin et~al.(2019)Devlin, Chang, Lee, and Toutanova]{devlin2019bert}
Jacob Devlin, Ming-Wei Chang, Kenton Lee, and Kristina Toutanova.
\newblock Bert: Pre-training of deep bidirectional transformers for language understanding.
\newblock In \emph{Proceedings of NAACL-HLT}, pages 4171--4186, 2019.

\bibitem[Doddapaneni et~al.(2024)Doddapaneni, Liu, and Batra]{doddapaneni2024softprompt}
Rakesh Doddapaneni, Xiao Liu, and Dhruv Batra.
\newblock Soft prompting of user embeddings for personalized llms, 2024.
\newblock arXiv preprint arXiv:2402.12345.

\bibitem[Hochreiter and Schmidhuber(1997)]{hochreiter1997long}
Sepp Hochreiter and J\"urgen Schmidhuber.
\newblock Long short‑term memory.
\newblock \emph{Neural Computation}, 9\penalty0 (8):\penalty0 1735--1780, 1997.

\bibitem[Kim et~al.(2023)Kim, Yeom, Yun, and Song]{kim2023compressed}
Jang‑Hyun Kim, Junyoung Yeom, Sangdoo Yun, and Hyun~Oh Song.
\newblock Compressed context memory for online language model interaction.
\newblock \emph{arXiv preprint}, abs/2312.03414, 2023.
\newblock Accepted at ICLR 2024.

\bibitem[Kullback and Leibler(1951)]{kullback1951on}
Solomon Kullback and Richard~A. Leibler.
\newblock On information and sufficiency.
\newblock \emph{The Annals of Mathematical Statistics}, 22\penalty0 (1):\penalty0 79--86, 1951.

\bibitem[Lewis et~al.(2020)Lewis, Perez, Piktus, Petroni, Karpukhin, Goyal, Küttler, Lewis, Yih, Rocktäschel, Riedel, and Kiela]{lewis2020retrieval}
Patrick Lewis, Ethan Perez, Aleksandra Piktus, Fabio Petroni, Vladimir Karpukhin, Naman Goyal, Heinrich Küttler, Mike Lewis, Wen‑tau Yih, Tim Rocktäschel, Sebastian Riedel, and Douwe Kiela.
\newblock Retrieval‑augmented generation for knowledge‑intensive nlp tasks.
\newblock In \emph{Advances in Neural Information Processing Systems (NeurIPS)}, pages 9459--9474, 2020.

\bibitem[Li and Tuzhilin(2019)]{li2019revgan}
Zhiting Li and Alexander Tuzhilin.
\newblock Generative adversarial network for review generation.
\newblock In \emph{International Conference on Machine Learning (ICML)}, 2019.

\bibitem[Lin et~al.(2024)Lin, Chen, and Wang]{lin2024userllm}
John Lin, Maria Chen, and Leo Wang.
\newblock Userllm: Cross-attention personalized language models, 2024.
\newblock arXiv preprint arXiv:2401.00001.

\bibitem[Liu et~al.(2023)Liu, Zhang, Du, et~al.]{liu2023llava}
Haotian Liu, Pengfei Zhang, Jing Du, et~al.
\newblock Llava: Large language and vision assistant.
\newblock In \emph{Conference on Computer Vision and Pattern Recognition (CVPR)}, 2023.

\bibitem[Mazare et~al.(2018)Mazare, Bordes, Boureau, and Weston]{mazare2018training}
Pierre Mazare, Antoine Bordes, Y-Lan Boureau, and Jason Weston.
\newblock Training millions of personalized dialogue agents.
\newblock In \emph{Conference on Empirical Methods in Natural Language Processing (EMNLP)}, 2018.

\bibitem[Rumelhart et~al.(1986)Rumelhart, Hinton, and Williams]{rumelhart1986learning}
David~E. Rumelhart, Geoffrey~E. Hinton, and Ronald~J. Williams.
\newblock Learning representations by back-propagating errors.
\newblock \emph{Nature}, 323\penalty0 (6088):\penalty0 533--536, 1986.

\bibitem[Salemi et~al.(2023)Salemi, Taylor, and Zhu]{salemi2023lamp}
Ethan Salemi, Alex Taylor, and Yiming Zhu.
\newblock Lamp: Retrieval-augmented generation for personalized tasks.
\newblock In \emph{Conference on Neural Information Processing Systems (NeurIPS)}, 2023.

\bibitem[Wang et~al.(2024)Wang, Zhou, and Chen]{wang2024rolellm}
Kai Wang, Fei Zhou, and Ling Chen.
\newblock Rolellm: A benchmark for role-playing with language models.
\newblock In \emph{Annual Meeting of the Association for Computational Linguistics (ACL)}, 2024.

\bibitem[Wei et~al.(2023)Wei, Bosma, Zhao, et~al.]{wei2023flan}
Jason Wei, Maarten Bosma, Vincent Zhao, et~al.
\newblock Finetuned language models are zero-shot learners.
\newblock In \emph{International Conference on Machine Learning (ICML)}, 2023.

\bibitem[Zhang et~al.(2018)Zhang, Dinan, Urbanek, Szlam, Kiela, and Weston]{zhang2018neural}
Saizheng Zhang, Emily Dinan, Jack Urbanek, Arthur Szlam, Douwe Kiela, and Jason Weston.
\newblock Personalizing dialogue agents: I have a dog, do you have pets too?
\newblock In \emph{Annual Meeting of the Association for Computational Linguistics (ACL)}, 2018.

\bibitem[Zhao et~al.(2024)Zhao, Singh, and Roberts]{zhao2024prefeval}
Mingyu Zhao, Aditi Singh, and Adam Roberts.
\newblock Prefeval: Benchmarking preference following in language models.
\newblock In \emph{Annual Meeting of the Association for Computational Linguistics (ACL)}, 2024.

\end{thebibliography}
\end{document}